\documentclass{article}
\usepackage[utf8]{inputenc}
\usepackage{graphicx}
\usepackage[colorlinks]{hyperref}
\usepackage{pifont}
\usepackage[numbers]{natbib}
\usepackage{authblk}

\title{A systematic literature review on Robotic Process Automation security} 
\graphicspath{Images/}
\begin{document}

\author [a]{Nishith Gajjar}
\author [b]{Keyur Rathod}
\author [c]{Khushali Jani}
\affil[a]{Engineering System and Computing, Graduate Student, University of Guelph, Ontario, Canada}
\affil[b]{Engineering System and Computing, Graduate Student, University of Guelph, Ontario, Canada}
\affil[c]{Engineering System and Computing, Graduate Student, University of Guelph, Ontario, Canada}

\date{September 2022}

\maketitle

\rule{\textwidth}{0.5pt}
\begin{abstract}
    The technocrat epoch is overflowing with new technologies and with such cutting-edge facilities accompanies the risks and pitfalls. Robotic process automation is another innovation that empowers the computerization of high-volume, manual, repeatable, everyday practice, rule based and un-motivating human errands. The principal objective of Robotic Process Automation is to supplant monotonous human errands with a virtual labor force or a computerized specialist playing out a similar work as the human laborer used to perform. This permits human laborers to zero in on troublesome undertakings and critical thinking. Robotic Process Automation instruments are viewed as straightforward and strong for explicit business process computerization. Robotic Process Automation comprises of an intelligence to decide if a process should occur. It has the capabilities to analyze the data presented and provide the decision based on the logic parameters set in place by the developer. Moreover, it does not demand for system integration, like other forms of automation. Be that as it may, since the innovation is yet arising, Robotic Process Automation faces a few difficulties during the execution.
    \end{abstract}


\section{Introduction}
Software called robotic process automation (RPA) makes it simple to create, use, and manage software robots that mimic how people interact with computers and software \cite{uipath-inc-no-date, a11}.In other words, Robotic process automation (RPA), an automation technology that enables organisations to partially or entirely automate routine processes, is controlled by business logic and structured inputs. Robotic process automation software robots, often known as "bots," can imitate human actions to execute tasks like data entering, transaction processing, reaction triggering, and interacting with other digital systems. Robotic Process Automation systems come in a variety of forms, ranging from straightforward online "chat bots" that can respond to common questions to massive deployments of thousands of bots that can automate tasks like credit card processing and fraud detection \cite{cyberark-software-2021}.
Robotic process automation is a procedure that uses the artificial intelligence and machine learning capabilities to handle the high-volume data task effectively. Distinct steps are included in Robotic Process Automation such as discovery, design, development, testing, and production or deployment. In the automation process, each phase has a prevailing impact. Finding the processes that can be automated is the goal of the discovery phase. To find the ideal candidate for automation, technical and business feasibility studies are conducted. The design phase includes creating the various process steps. Business analysts and solution architects, respectively, draught the process design document (PDD) and the solution design document(SDD). Then, bots based on process design document and Solution Design Document are being developed by developers. They even run unit tests to ensure that the development is proceeding properly during the development stage. The testing team can now use various test cases to do System Integration Testing (SIT) to test the BOT, and after it passes, either the testing team or the business team can undertake User Acceptance Testing (UAT). The code is deployed to the production environment once it has undergone testing and received approval from User Acceptance Testing (UAT) and System Integration Testing (SIT). The deployment phase is entered after the initial runs on the production environment. Robotic Process Automation uses tools in order to implement task like software application. It can be stated that Robotic Process Automation  is an error-free and risk-free process which can get more customer satisfaction and Return of Investment. On the financial and organizational perspective, it provides an aid in depleting the training cost, labour cost and boosting capabilities along with saving time. There are distinct sectors where Robotic Process Automation can work effectively like banking, human resources and Customer Relationship Management. Regardless of the perks, there are two main risks associated with Robotic Process Automation such as data leakage and fraud. Lacking in adequate security measures, the sensitive data that Robotic Process Automation bot credentials or customer data that Robotic Process Automation handles, can be exposed to attackers. To mitigate the security failures in Robotic Process Automation projects, security and risk management leaders need to follow a four-step action plan which consists of ensuring accountability for bot actions, avoiding abuse and fraud, protecting log integrity and enabling secure Robotic Process Automation development \cite{ a1, a2,unknown-author-2020} .

\subsection{Prior Research}
To our knowledge, relatively few Systematic Literature Reviews have been conducted specifically on the use of block chain to address the issue of cyber security (SLRs) \cite{a3}. In the area of robotic process automation in cyber security, one of the most current survey study was completed by \cite{agostinelli2019synthesis}
Robotic Process Automation cannot yet completely substitute human labour. Automation is limited to straightforward, predictable tasks. Whenever a specific situation arises for which the rule set does not provide an appropriate answer. Escalation to a human supervisor is possible thanks to robots. This is particularly true in modern application situations where even expected activities suddenly less predictable as a result of the massive amounts of data and events generated in these circumstances that could impact how they are implemented. 
Another study conducted for in-depth research on Robotic Process Automation is displayed in the book \cite{vaseashta2022applying} Instead of using robots to perform human jobs, robotic process automation uses software. Robotic Process Automation has recently gained popularity because it can automate repetitive and high-volume processes in order to reduce manual effort and errors and increase productivity. That is to say. By lowering errors, unsettling behaviour, conserving resources, and balancing variance, Robotic Process Automation promotes higher operational efficiency. This idea is adaptable and flexible. It facilitates seamless integration with already-in-place procedures and aids in lowering costs, maintaining quality, accelerating delivery times, and improving customer satisfaction. Robotic Process Automation, as it is used in practise, enables bots, or specifically created software programmes, to take over various complex processes and effectively carry out activities that are typically performed by humans. These include inventory and supply chain management, operational tasks, procure-to-pay processing, and data extraction and management.
However, just like any other technology, Robotic Process Automation has security problems and obstacles, some of which are shown in Figure 1. In order to ensure that there is no wrongdoing that could result in errors and harm, robots, or bots, must be constantly monitored at multiple levels. In addition, risk rises as the number of variables increases. Bots give malicious actors another attack vector when they are integrated into the system. This stands out in terms of data privacy, or the improper use of data. Cyber  criminals can, for example, use malicious software to gain unauthorised access to bot systems in order to steal sensitive user data and information. Additionally, since bots work quickly, they might be able to continue processing data in the event of a breach with a delayed response, even though they shouldn't. It may result in erroneous and faulty data. Although intelligent, bots are not infallible. Since they are not designed for intent identification, it may be difficult to identify a security compromise. The usage of Robotic Process Automation frameworks may expose enterprises to new kinds of online vulnerabilities. If Robotic Process Automation is not in compliance with regulations provided by regulatory organisations, their inclusion into operational and corporate activities may result in penalties. This cost of non-compliance results from the system being more complex.

\begin{center}
    \includegraphics[]{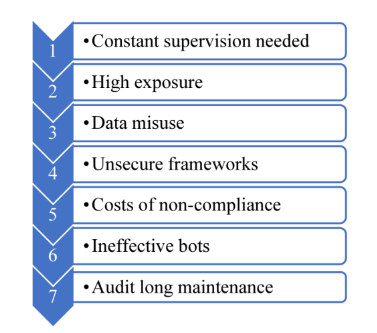}
    \center Figure-1
\end{center}

\subsection{Research Goals}
This study's objectives are to review prior research, summarise its conclusions, and concentrate on robotic process automation for Cyber security.

\subsection{Contributions and Layout}
In order to advance the work of people with an interest in robotic process automation and cyber security, this SLR complements existing research and offers the following additions:

\begin {table}[h!]
\caption{Research Questions}
\label{tab:table1}
    \begin{center}
        \begin{tabular}{p{5.5cm}    p{5.5cm}}
        \hline
        \textbf{Research Questions (RQ)} & \textbf{ Discussion }\\
        \hline
        \textbf{RQ(1)} What are the drawbacks of using Robotic Process Automation in cyber security? & Data theft, inadvertent use of privileged access, and denial-of-service are recurrent and enhancing Robotic Process Automation growth restrictions that place enterprises at serious risk.\cite{unknown-author-20212}\textit{}\\
        \\
        \textbf{RQ(2)} What can be considered as some of the best Robotic Process Automation practices to mitigate the risk in cyber security & Segregation of tasks is the primary security best practise. Digital identity and access is the second best practise for security. Data encryption is third on the list of security best practises. A further security best practise is to create policies for data classification, data retention, data storage, and data location.\cite{kosi2019robotic}\\
        \\
        \textbf{RQ(3)} What can be the best use of the Robotic Process Automation tools in cyber-security & Through the use of robotic process automation tools, firms can save spending hours each day on tiresome chores by automating repetitive processes. Custom scripts that automate these operations can also be written using the Robotic Process Automation technologies.\cite{malak-2022}\\
        \\
        
        \hline
        \end{tabular}
    \end{center}
\end{table}

\begin{itemize}
    \item Up until mid-2022, we identified 35 primary research publications and documents connected to robotic process automation. This SLR may be cited by other researchers to further their work.
    \item We narrowed down these 35 research papers that were chosen to 23 that perfectly matches this SLR. These articles can be consulted for advice on any inference.
    \item These 23 research articles and materials served as the basis for a thorough examination that led to the thoughts, deductions, and conclusions we reached on the subject of robotic process automation in cyber security.
    \item By adhering to these documents, we suggest developing a standard to offer assistance in any research projects including robotic process automation in cyber security.
\end{itemize}
The format of this article is as follows: The techniques used to choose the primary studies for analysis in a methodical manner are described in \hyperref[sec:ResMeth]{Section 2}.
The results of all the primary research chosen are presented in \hyperref[sec:fndgs]{Section 3}.\hyperref[sec:discs]{Section 4} addresses the conclusions in relation to the earlier-presented study questions.\hyperref[sec:conclandfut]{Section 5} wraps up the study and makes some recommendations for additional research.
\section{Research Methodology}
\label{sec:ResMeth}
To accomplish the goal of addressing the exploration questions, we
led the Systematic Literature Review. We looked to travel through the preparation, directing, and detailing periods of the survey in emphasis to consider a careful assessment of the Systematic Literature Review.
\subsection{Selection of Primary Studies}
\noindent Essential investigations were featured by passing watchwords to the inquiry office of a specific distribution or web index. The catchphrases were chosen to advance the rise of examination results that would help with responding to the exploration questions.\\
The platform that were used to make the search are:
\begin{enumerate}
    \item Google Scholar
    \item Science Direct
    \item SpringerLink
    \item Association for Computing Machinery
    \item IEEE Xplore digital Library
    \item arXiv e-Print Archive
    \item Research Gate
    \item Social Science Research Network(SSRN) \cite{8501753}  \cite{10.1007/978-981-16-3961-6_20}.
\end{enumerate}
\subsection{Inclusion and Exclusion Criteria}
\noindent Studies to be remembered for this Systematic Literature Review should report observational discoveries and could be papers on contextual investigations, new specialized blockchain applications also, and critiques on the improvement of existing security components through blockchain coordination \cite{a4,s5}. They should be peer-checked and written in English. Any outcomes from Google Researcher will be checked for consistency with these measures as there is an opportunity for Google Researcher to return lower-grade papers. Just the latest variant of a review will be remembered for this Systematic Literature Review.The key inclusion and exclusion are depicted in the \hyperref[tab:table2]{Table 2}.
\begin{table}[h!]
    \caption{Inclusion and Exclusion criteria}
    \label{tab:table2}
    \begin{center}
    \begin{tabular}{p{5cm}  p{6cm}}
    \hline
    \textbf{Standards for consideration} & \textbf{Rules for avoidance}\\
    \hline The paper should introduce observational information connected with the application and the utilization of Robotic process Automation security & Papers zeroing in on financial, business or on the other hand legitimate effects of Robotic Process Automation security applications.\\
    \hline The paper should contain data connected with Robotic Process Automation security or related appropriated record advancements  & writings like websites and government archive.\\
    \hline The paper should be a peer-reviewed surveyed item distributed in a gathering continuing or journal & No-English paper\\
    \hline
\end{tabular}
    \end{center}
\end{table}
\newpage
\subsection{Selection Results}
\noindent There was a sum of 187 investigations recognized from the underlying watchword look through the chosen stages. This was diminished to 160 later eliminating copy studies. In the wake of really looking at the examinations under the consideration/rejection rules, the number of papers staying for perusing was 25. The 25 papers were perused in full with the consideration/prohibition rules being re-applied, and 20 papers remained. Forward and reverse compounding distinguished an extra 5 and 5 papers separately, giving the last figure for the number of papers to be remembered for this Systematic Literature Review as 22.
\subsection{Quality Assessment}
\noindent An appraisal of the nature of essential investigations was made concurring with the direction. This took into consideration an appraisal of the significance of the papers to the examination questions, with thought to any indications of exploration predisposition and the legitimacy of trial information. Five arbitrarily chosen papers were exposed to the accompanying quality evaluation cycle to really take a look at their viability\cite{alfandirobotic}:\\
Stage-1: \textbf{Robotic Process Automation}. The paper should be centered around the utilization of Robotic Process Automation or the utilization of Robotic Process Automation innovation to a particular an issue very much remarked.\\
Stage-2: \textbf{Context}. Enough setting should be accommodated in the exploration
goals and discoveries. This will take into consideration an exact understanding of the exploration.\\
Stage-3 \textbf{Robotic Process Automation application}. There should be an adequate number of subtleties in the study to make a precise show of how the innovation has been applied to a particular issue.\\
Stage-4 \textbf{Security factors}. The paper should give clarification to the security issue.\\
Stage-5 \textbf{Robotic Process Automation Performance}. Surveying the presentation of
Robotic Process Automation in the climate for which it is applied will permit for correlations of various Robotic Process Automation applications.\\
Stage-6 \textbf{Data Gathering}. Insights regarding how the information was procured, estimated and announced should be given to deciding exactness.\\ 
\cite{unknown-author-2021}
\subsection{Data Extraction}
The information extraction should be applied to all finding that has finished the quality evaluation assessment that can be seen in the  \hyperref[fig:figure2]{Figure2}. Be that as it may, to check regardless of whether the strategy of information extraction is appropriate, we applied this cycle to the underlying five discoveries. Once, we come by the ideal outcome, the information extraction process is applied to all articles.\\
\textbf{Context Data}: The information about the point and extreme objective of the review.\\
\textbf{Qualitative Data}: The records and inductions are given by creators and looked into by peers.\\
\textbf{Quantitative Data}:The information gathered in light of any measurements or any exploratory examinations.\\
\subsection{Data Analysis}
To meet the goal of addressing the exploration questions, we incorporated the information held inside the subjective and quantitative information classes. Furthermore, we directed a meta-examination of those papers that were exposed to the last information extraction process.
\begin{flushleft}

\begin{center}
    \includegraphics[]{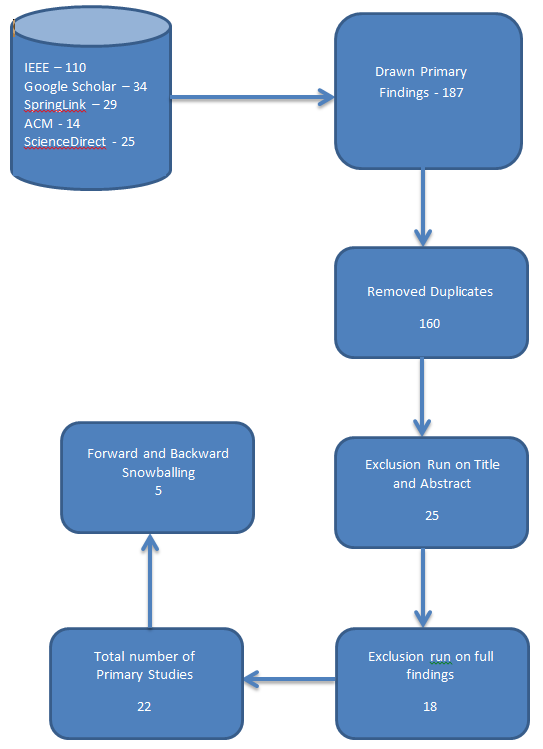}
    \label{fig:figure2}
    \center Figure-2
\end{center}

\begin{center}
    \includegraphics[]{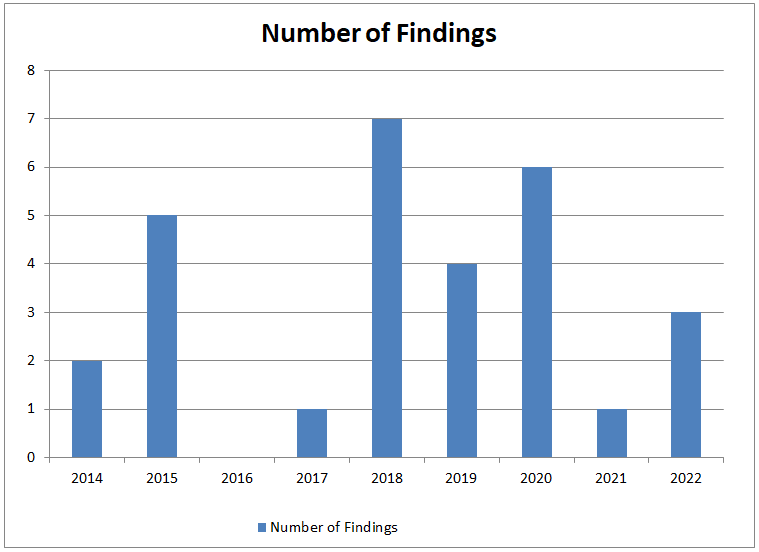}
    \label{fig:figure3}
    \center Figure-3
\end{center}

\subsubsection{Significant word Counts}
To sum up, the normal subjects among the chosen
essential investigations, an examination of watchwords was performed across each of the 22
studies.\hyperref[tab:table3]{Table 3} shows the times a few explicit words showed up in the essential examinations in general. As should be visible in the table, barring the watchwords chosen by the creator, i.e., "Robotic Process Automation" and "security", the third catchphrase showing up most often in our dataset is "Robotic Process Automation", later "Security" and "Bots". This shows rising interest in the reception of Robotic Process Automation with regard to security.

\begin {table}[h!]
\caption{Keyword frequencies in Primary Studies }
\label{tab:table3}
\begin{center}
\begin{tabular}{p{5 cm}    p{2 cm}}
        \hline
        \textbf{Keywords} & \textbf{Count}\\
        \hline Automation & 1196\\
        \hline Process & 1372\\
        \hline Bots & 688\\
        \hline Robotic & 1035\\
        \hline Security & 756\\
        \hline Network & 159\\
        \hline Attacks & 420\\
        \hline Risk & 942\\
        \hline Cyber & 853\\
        \hline Authentication & 208\\
        \hline Information & 116\\
        \hline Detection & 137\\
        \hline Privacy & 206\\
        \hline Access & 186\\
        \hline 
\end{tabular}
    \end{center}
\end{table}
\newpage
The aforementioned points can be considered as some of the major pitfalls in the field of Robotic Process Automation in cyber security. Therefore, prominent solutions can be applied to mitigate the concerned issues which can in turn provide more secured data and better security. This research paper will mainly focus on finding the considerable solutions for the proposed complications that can be a hindrance in securing the data that are really crucial for the organizations. 
\end{flushleft}
\section{Findings}
\label{sec:fndgs}
Each paper's ideas and pertinent research have been examined and are provided in These studies were conducted with an emphasis towards common threats \cite{a6,a7}, issues with Robotic Process Automation, and potential fixes that might be provided using cyber security or other cutting-edge emerging technologies. 

Additionally, each document was divided into a distinct category to make the analysis simpler. For instance, initial research on Robotic Process Automation security is included under security. Similar to that, a study that focuses on Robotic Process Automation application falls under the heading of Robotic Process Automation application. Additionally, the study focuses on employing various techniques to find a defence against various Robotic Process Automation attacks. Thus, it is evident that approximately (48\%) of the documents are concerned with the security of robotic process automation. Additionally, for Robotic Process Automation solutions and Robotic Process Automation development, we received (28\%) and (32\%) respectively.

The provided \hyperref[tab:table4]{Table4} shows the statistics related to these discoveries. The research comprises a study of Robotic Process Automation security, an examination of the dangers and attacks that affect Robotic Process Automation security, and any workable conclusions that might help to strengthen security. Studies that concentrate on the applications of Robotic Process Automation include information on the need for automation, challenges that can be solved with Robotic Process Automation, how network security will be boosted by Robotic Process Automation, and other topics. On the other hand, if cutting-edge and new technology, like machine learning, can be employed to solve the problem, it can be handled better.
\section{Discussions}
\label{sec:discs}
Initial keyword searches reveal that there are a considerable number of publications that are connected to Robotic Process Automation. Robotic Process Automation technology and truly distributed, decentralised systems have just recently been invented and are still in their infancy. The chosen main studies contain a significant proportion of experimental hypotheses or notions with limited quantitative information and few applications to real-world situations. The initial discoveries were many, focused on what Robotic Process Automation is, why Robotic Process Automation is required, and what issues Robotic Process Automation can resolve. This was done while searching with the security of Robotic Process Automation.
Due to its ability to accelerate business growth by eliminating a significant amount of manual and repetitive work, Robotic Process Automation has recently attracted a lot of interest from both the commercial and academic worlds. But there are still a lot of difficulties with Robotic Process Automation implementation right now. The inability to analyse process priorities (40\%),absence of risk management tools (28\%), insufficient internal staff skills (24\%), and lack of urgency (23\%), among others, are issues at the organisational structure level, according to the Global Robotic Process Automation Survey 2019 study\cite{unknown-author-no-date(c)}, which is represented in \hyperref[fig:figure4]{Figure 4}. Information and data security (40\% of the technical risk), scaling challenges (37\%), and choosing an appropriate development platform (30\%). Inappropriate application scenarios (32\%), increased implementation costs (37\%), and external legal and regulatory constraints (30\%) are some of the financial and regulatory factors. A further discussion on these challenges is presented in \hyperref[fig:figure5]{Figure 5}\cite{inbook}.

\begin{center}
    \includegraphics[]{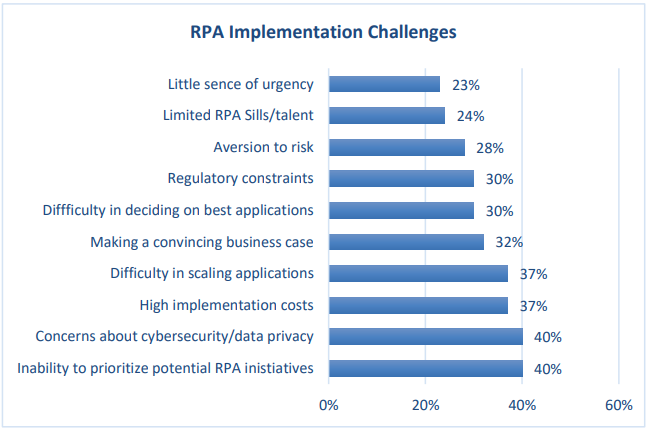}
    \center Figure-4
        \label{fig:figure4}
\end{center}
\newpage
\begin{center}
    \includegraphics[]{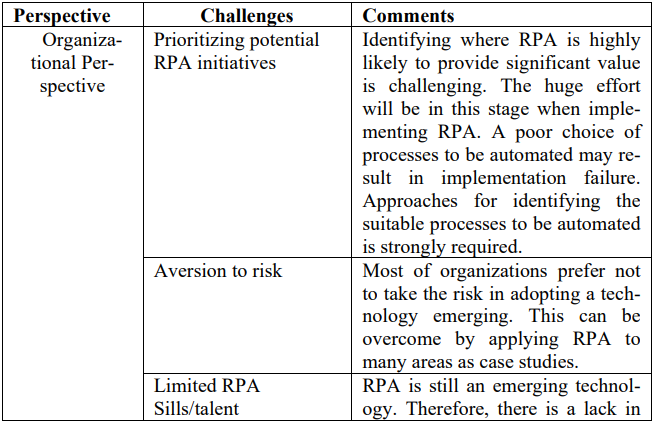}
        \includegraphics[]{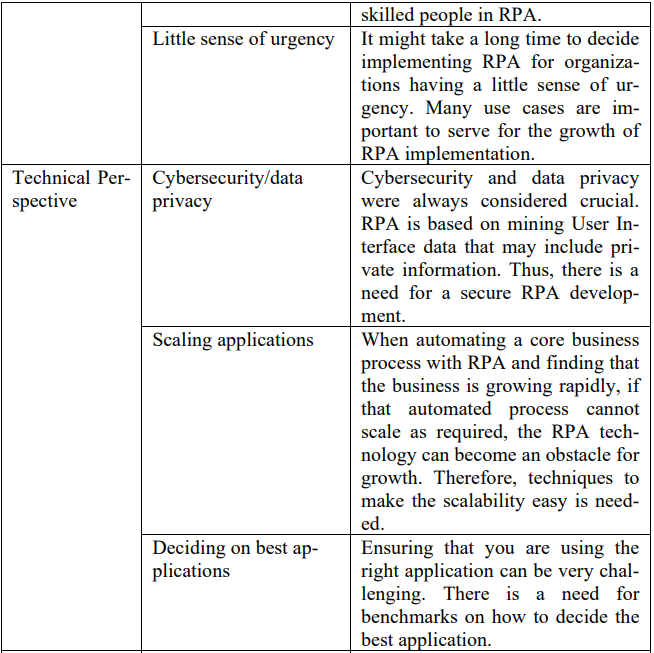}
    \includegraphics[]{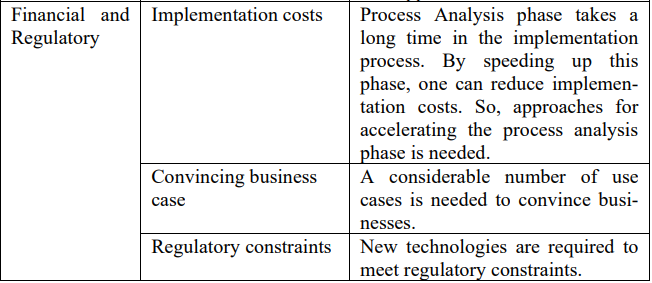}
    \center Figure-5
        \label{fig:figure5}
\end{center}

\subsection{RQ1:What are the drawbacks of using Robotic Process Automation in cyber security? }
It is crucial to emphasise that the purpose of this systematic literature review is to find solutions to the problems caused by Robotic Process Automation's use in cyber security, not to focus on its benefits. Robotic process automation (RPA) adoption has many implications. However, it also faces a myriad of challenges, such as cyber threats. Data theft, abuse of privileged access, and denial-of-service are frequent and developing Robotic Process Automation growth restrictions that present serious threats to businesses.

Robotic Process Automation carries security vulnerabilities that constitute careful planning for prevention. Because of the vulnerability of the technology to cyberattacks, organisations are at risk. It is important for a robotic automation system to take into account both the business process and the security concerns because firms are gradually embracing Robotic Process Automation-based technology as a digital strategy for business development. This will allow for the implementation of safety measures and checks.

Since Robotic Process Automation credentials are frequently exchanged and utilised, the first thing that needs to be ensured is that they are not left unmodified and unprotected. This could expose the system to a future Cyber attack wherein the passwords are stolen and utilized for malicious reasons\cite{unknown-author-20212}.
However, in order to maximise their technological investments, business leaders must also comprehend and evaluate the possible hazards of Robotic Process Automation. Although Robotic Process Automation can drive innovation and optimise competitiveness, organisations frequently establish unreasonable goals and expectations for Robotic Process Automation implementation, or misuse it for a one-off, isolated area. These factors can lower profitability, damage employee productivity, and disrupt company workflows. As a result, any Robotic Process Automation efforts suffer from under-resourcing because Robotic Process Automation is unable to live up to its promise of delivering greater value. Organizations that just use Robotic Process Automation to reduce costs by lowering FTE manpower rather than utilising it to innovate and improve how work is done, lack any strategic goal or end-point design in their Robotic Process Automation projects. A sound, future-proof target operating model must be put in place, and the appropriate intelligent process automation tools must be used to reduce the risk associated with Robotic Process Automation. approach\cite{kiran-2022}. 
\subsection{RQ2:What can be considered as some of the best Robotic Process Automation practices to mitigate the risk in cyber security?}
Some Robotic Process Automation security best practices are summarised below.
\begin{itemize}
\item Choose Robotic Process Automation carefully.

Robotic Process Automation developers vary greatly from one another. Information security needs to be taken into account along with functional specifications when choosing a new Robotic Process Automation technology. Malicious code or security flaws could be present in a bot with inadequate coding\cite{unknown-author-2021}.
\item Create a security governance framework for Robotic Process Automation.

Regular risk analyses and audits of Robotic Process Automation processing activities must be part of an Robotic Process Automation governance structure. Employees in charge of the Robotic Process Automation must have a comprehensive understanding of their security obligations, which include restricting access to the environment, logging and tracking activity, and more\cite{unknown-author-2021(b)}.

\item Avoid using hard-coded access rights.

Robot scripts must replace all hard-coded access permissions with API calls, with each request linked directly to the required access privileges kept in a single repository. An additional layer of defence is added, decreasing the likelihood of an attack \cite{kaur-2022}.

\item Build in-error handling

Automation can be halted by errors like unsuccessful login attempts, missing directories, or running out of disc space. Automation may also be slowed down by glitches like a timed-out application, inaccurate data, or a new screen inside the application. Workflows should include error handling for this reason. An Robotic Process Automation programmer should programme the automation to manage the exception and respond appropriately depending on the type of exception that happens, whether it is a business or application exception. For instance, the Robotic Process Automation bot should log the business error and set up the environment to handle queue item number three if it happens on item number two in the queue. The bot should bounce back from errors and keep working through all the transactions\cite{unknown-author-2022}.
\end{itemize}
\subsection{RQ3:What can be the best use of the Robotic Process Automation tools in cyber-security?}
Based on the features they offer and the feedback from users, the following manufacturers offer some of the best Robotic Process Automation solutions available on the market.
\begin{itemize}
    \item UiPath
    
With the aim of streamlining, accelerating, and optimising digital transformation for businesses, UiPath is an amazing and user-friendly Robotic Process Automation platform that enables users to automate their manual operations fast and efficiently.

By automatically analysing a company's operations, UiPath can decide which ones should be automated. In addition to automating mundane tasks like data entry, email marketing, and site scraping, it also takes care of recurring obligations like notice, documentation, and set-up follow-ups. UiPath provides capabilities like encryption and role-based access control in addition to automation that is simple to set up. It can also manage processes of any size or complexity\cite{fulmer-2022}.

\item Blue Prism

An Robotic Process Automation tool called Blue Prism has the ability to create a software-powered virtual workforce. This enables businesses to automate business processes in a flexible and economical way. The application features a visual designer with drag-and-drop functionality and is based on the Java programming language.
Blue Prism provides a visual designer that is free of any interference, recorders, or scripts \cite{kappagantula-2022}.

\item Kofax

The workflows are referred to as robots in Kofax Robotic Process Automation. You are free to explore the applications as you combine them while building a robot. You can log into programmes, extract data from pages, fill out forms or search boxes with information, choose options from menus, and scroll through numerous pages. Additionally, your robot has access to databases, files, web services, APIs, and other robots. It can export data from one application and load it into another, altering it as necessary in the process.
You can automate Windows and Java programmes on your network devices with the help of Kofax Robotic Process Automation's Desktop Automation feature. Desktop automation replaces manual operations by invoking a desktop or terminal application \cite{unknown-author-no-date(d)}.

\item Pega Robotic Process Automation

Organizations may automate those tiresome, time-consuming manual operations using Pega Robotic Process Automation (RPA). Robotics connects old systems and gets away of tedious manual data entry. By converting manual processes to digital ones, low-value operations become much more predictable, allowing workers to concentrate on more strategic responsibilities. Businesses may use bots that produce dependable outcomes by using the power of Pega Robotic Process Automation. Using a visual interface, workflows can be rapidly and effectively drawn out and updated as your organisation grows \cite{unknown-author-2021(a)}.

\end{itemize}
\section{Future research directions of Robotic Process Automation }
We provide the following research directions for robotic process automation for cyber security that need additional study based on the findings of this survey and our observations:
\begin{itemize}
    \item \textbf {Integration of additional tools and SPA}
    
Artificial intelligence and machine learning are becoming a part of Robotic Process Automation. We may anticipate that in the near future, Robotic Process Automation will support both straightforward judgment-based automation and the processing of unstructured data. This will assist Robotic Process Automation in moving past rule-based technology.
Robotic Process Automation will increasingly be integrated with other tools and technologies as businesses embrace it to automate their activities. To improve the features and simplify automation, other tools will be incorporated with it. Smart Process Automation is the abbreviation. Robotic Process Automation is currently having some difficulty automating the process of handling unstructured data. The unstructured data process will be automated with the aid of SPA, which combines a number of various technologies like machine learning, AI, and cloud technology\cite{a7,a8, unknown-author-no-date}.

\item \textbf {RPA's effective evolution} 

Robotic Process Automation will eventually be able to recognise and enhance processes within and across your systems without the need for human interaction as a result of technological improvements. In other words, your company will be able to completely get in front of processes rather than just automate them. Process management and Robotic Process Automation will soon be used interchangeably. The automation perspective will be applied to every business function. Leading analysts forecast that Robotic Process Automation will soon become a common tool for boosting productivity. The Robotic Process Automation tool's ability to work alongside intelligent enterprise automation, a group of integrated technologies that may include intelligent capture, artificial intelligence, machine learning, case management, workflow, low-code capabilities, and cloud-based content services, will be a key differentiator\cite{payne-2022,a9.a11}.

\item \textbf {Boosting security with RPA} 

Software robots that automate manual tasks can increase productivity, decrease errors, increase income, and provide a host of other advantages for businesses. But the use of Robotic Process Automation in cybersecurity is one of its most attractive and significant applications. By automating many of the manual processes that these professionals still utilise, Robotic Process Automation may have a big positive impact while enabling them to contribute their own expertise and insight when it matters most. Robotic Process Automation may, of course, offer a significant level of automation to the overall cybersecurity workflow, but it's also critical to make sure the Robotic Process Automation platform is safe. The platform should also work well with other security measures already in place, such as user authentication and permission systems, to guarantee the security of any manual activities it automates\cite{davis-2021.a9}.
\end{itemize}

\section{Conclusion and Future Work}
\label{sec:conclandfut}
When implementing RPA, security must be considered; it cannot be added as a "bolt-on" feature later. In a summary, meticulous planning should go into RPA installation. This includes choosing a software vendor or platform that is well-known and has the security features previously mentioned, as well as implementing or incorporating your RPA users in corporate governance and security protocols. Constant oversight to guarantee compliance\cite{lewis--2019}.

By institutionalising crucial security features like identity verification, access control, data encryption, deployment security, and bot monitoring, one may leverage critical automation to help any organisation save money and become more productive while maintaining security\cite{dunlap-2022}.

RPA will be crucial in the future for creating a seamless organisational context because it has the ability to lower errors and boost efficiency. Repetitive jobs will be finished more quickly and efficiently, allowing people to focus on abilities that are more human-centric, such reasoning, judgement, and emotional intelligence\cite{capacity-2021,a11,a5}.

\section{Declarations of interest}
None.

\newpage
\begin{table}[h!]
\caption{Findings and Themes of the primary studies\cite{prokopets-2022}}.
\label{tab:table4}
\begin{center}
\begin{tabular}{p{2 cm} p{6 cm} p{4 cm}}
\hline\textbf{Primary Study} & \textbf{Qualitative and Quantitative Key Data Reported} & \textbf{Application Category}\\
\hline

\hyperlink{[S1]}{[S1]} & Periodic software refreshes are vital for network protection since they fix up security openings and programming holes in web applications and eliminate all weaknesses. & Robotic process automation bots\\

\hyperlink{[S2]}{[S2]} & Robotic Process Automation bots can computerize numerous information-related assignments to produce network safety alarms & Robotic Process Automation\\

\hyperlink{[S3]}{[S3]} & With Robotic Process Automation bots running explicit undertakings, unapproved clients are consequently kept from getting to your association's delicate or confidential information. & Robotic Process Automation Bots.\\

\hyperlink{[S4]}{[S4]} & From sending delicate data to the wrong email locations to misconfiguring resources for taking into account undesirable admittance to erroneously distributing private information on open sites, numerous associations have endured in view of carelessness on a piece of their HR. & Robotic Process Automation Bots.\\

\hyperlink{[S5]}{[S5]} & Digital danger hunting alludes to the course of persistently looking and filtering organizations to recognize and detach progressed digital dangers. & Artificial Intelligence enabled Robotic Process Automation bots\\

\hyperlink{[S6]}{[S6]} & Cyber Security examiners can use Robotic Process Automation bots to naturally convey security controls at whatever point weaknesses or irregularities are distinguished in frameworks.& Robotic Process Automation bots\\

\hyperlink{[S7]}{[S7]} & The robot/bot has ill-advised or wide admittance to frameworks/applications. The risk is somebody breaks the bot's certifications and can utilize the frameworks the bot approaches as a highlight assault. & Robotic Process Automation Security.\\

\hyperlink{[S8]}{[S8]} & Focuses on how a paradigm shift in the conception of software robots that are able to operate intelligently and flexibly in numerous dynamic and knowledge-intensive situations that are typical in today's application scenarios might tackle the problem of threats in Robotic process automation. & Robotic Process Automation and bots \\
\end{tabular}
\end{center}
\end{table}

\newpage
\begin{table}[h!]
\caption{Findings and Themes of the primary studies\cite{prokopets-2022}}.
\label{tab:table4}
\begin{center}
\begin{tabular}{p{2 cm} p{6 cm} p{4 cm}}
\hline\textbf{Primary Study} & \textbf{Qualitative and Quantitative Key Data Reported} & \textbf{Application Category}\\

\hyperlink{[S9]}{[S9]} & This article puts forth a multisectoral, interdisciplinary, local, state, and overall government systems strategy that combines hybrid challenges to its social, infrastructural, and informational dependencies. & Robotic Process Automation and security \\

\hyperlink{[S10]}{[S10]} & This study adds to the body of knowledge in the field of intelligent automation in banking and sheds light on its development and application. & Robotic Process Automation and banking sector security \\

\hyperlink{[S11]}{[S11]} & The cybersecurity implications of Robotic Process Automation technology are covered in this thesis, along with a risk analysis and a study of security best practises.& Robotic Process Automation and risk analysis \\

\hyperlink{[S12]}{[S12]} & This essay discusses the history and development of robotic process automation (RPA), as well as the distinctions and areas of expertise of the top Robotic Process Automation businesses and the various orchestrator technologies they employ. & Robotic Process Automation and business applications. \\

\hyperlink{[S13]}{[S13]} & The threat model discussed is intended on how to evaluate the effectiveness and security of robotic systems that interact with other machines while taking into account threats and vulnerabilities particular to such systems. & Robotic Process Automation and secuirty \\

\hyperlink{[S14]}{[S14]} &  Examined a number of security issues related to bot automation and offered a suggestion for how to create a comprehensive security framework for the Robotic Process Automation environment. & Robotic process automation and framework secuirty. \\

\hyperlink{[S15]}{[S15]} & The parallel ensemble model for threat hunting presented in this research is based on IIoT edge device behaviour abnormalities. Robotic Process Automation in IIOT edge device. \\

\hyperlink{[S16]}{[S16]} & Even while Robotic Process Automation implementation offers significant quantitative advantages, only a small number of businesses in each industry disclosed their quantitative outcomes.& Robotic Process Automation in small sector. \\
\end{tabular}
\end{center}
\end{table}

\newpage
\begin{table}[h!]
\caption{Findings and Themes of the primary studies\cite{prokopets-2022}}.
\label{tab:table4}
\begin{center}
\begin{tabular}{p{2 cm} p{6 cm} p{4 cm}}
\hline\textbf{Primary Study} & \textbf{Qualitative and Quantitative Key Data Reported} & \textbf{Application Category}\\
\hyperlink{[S17]}{[S17]} & Our research results in the creation of a process model that explains how companies respond to issues between IS security compliance and digital transformation by moving from preventive to reaction.& Robotic Process Automation and IS security \\

\hyperlink{[S18]}{[S18]} & The proposed design seeks to have a favourable effect on the Robotic Process Automation in terms of improved security and decreased latency. & Robotic Process Automation and its security. \\

\hyperlink{[S19]}{[S19]} & AI and intelligent automation are essential for businesses wishing to advance their Robotic Process Automation, and this article explains how to offer those solutions. & Robotic Process Automation and its security  \\

\hyperlink{[S20]}{[S20]} & This article gives a general summary of Robotic Process Automation and how to resolve its difficulties. & Difficulties of Robotic Process Automation \\

\hyperlink{[S21]}{[S21]} & In order to maximise their technological investments, business leaders must also comprehend and evaluate the possible hazards of Robotic Process Automation and this article shows how to perform the same. & hazards of Robotic Process Automation \\

\hyperlink{[S22]}{[S22]} & The possible security issues associated with Robotic Process Automation, go over secure Robotic Process Automation best practises, and provide a step-by-step action plan. & Security Issues of Robotic Process Automation\\

\hyperlink{[S23]}{[S23]} & Information about a company's employees, clients, and suppliers is accessible to the automation platform. & Information of Process Automation \\

\hyperlink{[S24]}{[S24]} & A list of Robotic Process Automation programming best practises that should be followed by developers with energy and flexibility. & Practice of Robotic Process Automation\\

\hyperlink{[S25]}{[S25]} & Artificial intelligence (AI) and machine learning (ML) are used in robotic process automation (RPA) software to learn and enhance manual procedures while reducing the workload of human personnel. & Importance of AI and ML in Robotic Process Automation \\

\hyperlink{[S26]}{[S26]} & The application features a visual designer with drag-and-drop functionality and is based on the Java programming language. & Functionality of Process Automation\\

\end{tabular}
\end{center}
\end{table}

\newpage
\clearpage
\bibliographystyle{ieeetr}
\bibliography{ref1}

\end{document}